\documentclass[letterpaper]{article} 
\usepackage{aaai2026}  
\usepackage{times}  
\usepackage{helvet}  
\usepackage{courier}  
\usepackage[hyphens]{url}  
\usepackage{graphicx} 
\usepackage{amsmath}
\usepackage{booktabs}
\usepackage{amssymb}
\usepackage{multirow}
\usepackage{xcolor}
\usepackage{subcaption}
\usepackage{natbib}  
\usepackage{caption} 
\usepackage{algorithm}
\usepackage{algorithmic}
\usepackage{pifont}
\usepackage{newfloat}
\usepackage{listings}

\newcommand{\Encoder}{\mathcal{M}}
\newcommand{\Feature}[2]{F^{#1}_{#2}}

\newcommand{\feature}[3]{{#1}^{#2}_{#3}}
\newcommand{\featureT}[3]{\tilde{#1}^{#2}_{#3}}

\newcommand{\Loss}{\mathcal{L}}

\newcommand{\Prm}[2]{P^{#1}_{#2}}

\DeclareCaptionStyle{ruled}{labelfont=normalfont,labelsep=colon,strut=off} 
\floatstyle{ruled}
\newfloat{listing}{tb}{lst}{}
\floatname{listing}{Listing}
\title{Hierarchical Prompt Learning for Image- and Text-Based Person Re-Identification}
\author {
    Linhan Zhou\textsuperscript{\rm 1}\equalcontrib,
    Shuang Li\textsuperscript{\rm 2}\equalcontrib,
    Neng Dong\textsuperscript{\rm 3},
    Yonghang Tai\textsuperscript{\rm 4},
    Yafei Zhang\textsuperscript{\rm 1},
    Huafeng Li\textsuperscript{\rm 1}\thanks{Corresponding author}
}
\affiliations{
    \textsuperscript{\rm 1}Faculty of Information Engineering and Automation, Kunming University of Science and Technology \\
    \textsuperscript{\rm 2}School of Computer Science and Technology, Chongqing University of Post and Telecommunication \\
    \textsuperscript{\rm 3}School of Computer Science and Engineering, Nanjing University of Science and Technology \\
    \textsuperscript{\rm 4}School of Physics and Electronic Information, Yunnan Normal University \\
    LinhanZhouUltra@outlook.com, shuangli936@gmail.com, neng.dong@njust.edu.cn,
    taiyonghang@126.com,
    
    \{zyfeimail, hfchina99\}@163.com
}

\usepackage{bibentry}

\begin{document}

\maketitle

\begin{abstract}
Person re-identification (ReID) aims to retrieve target pedestrian images given either visual queries (image-to-image, I2I)  or textual descriptions (text-to-image, T2I). Although both tasks share a common retrieval objective, they pose distinct challenges: I2I emphasizes discriminative identity learning, while T2I requires accurate cross-modal semantic alignment. Existing methods often treat these tasks separately, which may lead to representation entanglement and suboptimal performance.
To address this, we propose a unified framework named Hierarchical Prompt Learning (HPL), which leverages task-aware prompt modeling to jointly optimize both tasks. Specifically, we first introduce a Task-Routed Transformer, which incorporates dual classification tokens into a shared visual encoder to route features for I2I and T2I branches respectively. On top of this, we develop a hierarchical prompt generation scheme that integrates identity-level learnable tokens with instance-level pseudo-text tokens. These pseudo-tokens are derived from image or text features via modality-specific inversion networks, injecting fine-grained, instance-specific semantics into the prompts. Furthermore, we propose a Cross-Modal Prompt Regularization strategy to enforce semantic alignment in the prompt token space, ensuring that pseudo-prompts preserve source-modality characteristics while enhancing cross-modal transferability.
Extensive experiments on multiple ReID benchmarks validate the effectiveness of our method, achieving state-of-the-art performance on both I2I and T2I tasks.
\end{abstract}

\begin{links}
	\link{Code}{https://github.com/LH-Z-Ac/HPL-AAAI26}
\end{links}

\section{Introduction}

Person Re-Identification (ReID) aims to retrieve a target individual from large-scale visual data given a query, and plays a key role in surveillance and public security. Based on query modality, ReID is divided into Image-to-Image (I2I) and Text-to-Image (T2I)\cite{T2I1} tasks. I2I focuses on extracting identity-discriminative features robust to viewpoint and background changes, while T2I retrieves pedestrian images using natural language descriptions, enabling applications without image queries or requiring human interaction.
Recent advancements have led to the development of CNN- and Vision-Transformer\cite{ViT, MVC1, MVC2}-based models for I2I \cite{I2I-Conv1, I2I-Conv2, I2I-ViT1, I2I-ViT2, I2I-ViT3}, and fine-grained cross-modal alignment methods for T2I \cite{T2I-FineGrain1, T2I-FineGrain2, T2I-FineGrain3, T2I-FineGrain4, T2I-FineGrain5}. However, most existing works treat these tasks in isolation, neglecting the heterogeneous nature of real-world applications where both image and text queries may co-exist. This motivates the urgent need for a unified framework that can simultaneously handle both I2I and T2I ReID tasks flexibly.

\begin{figure}[t]
    \centering
    \includegraphics[width=\linewidth]{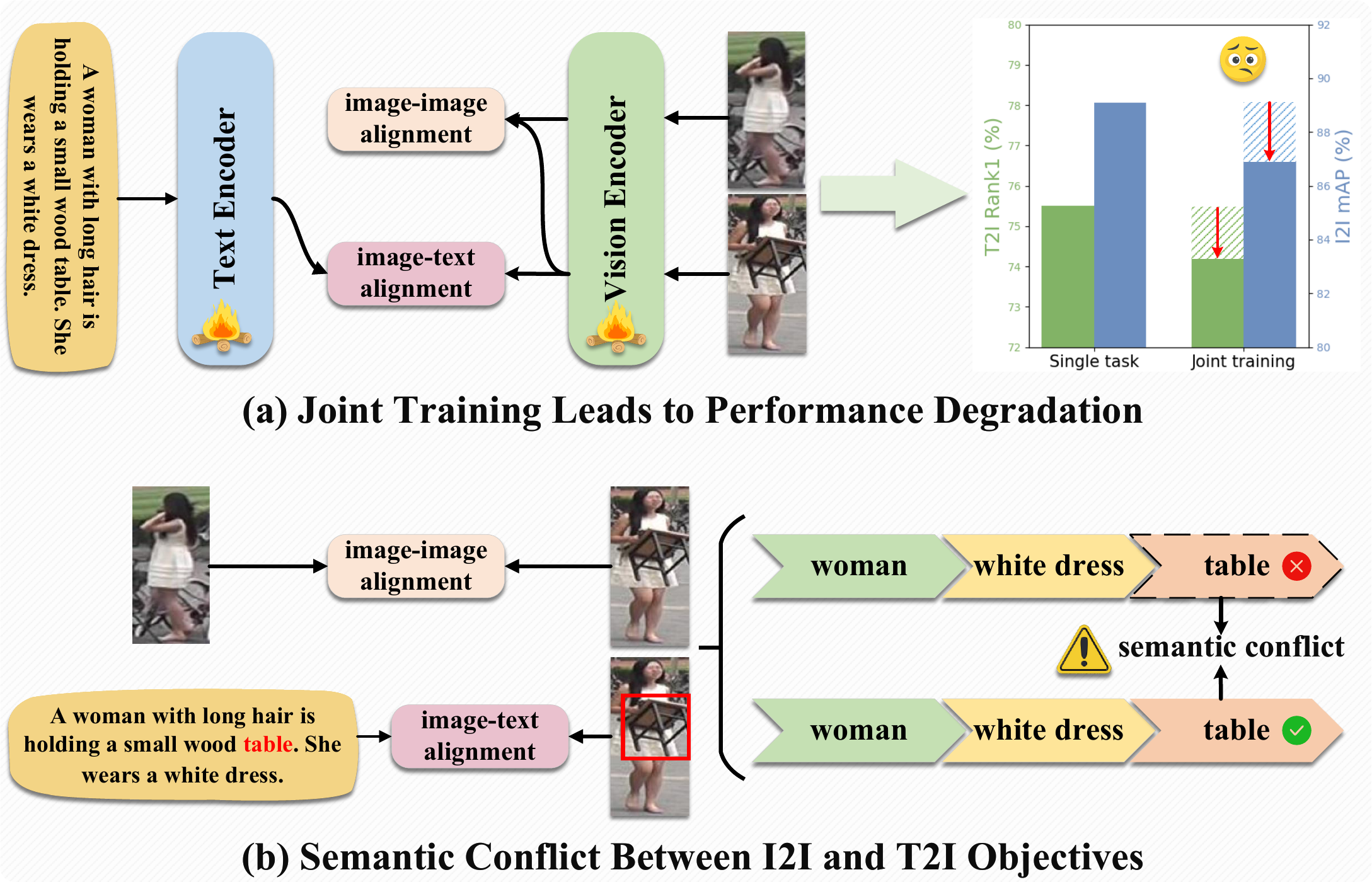}
    \caption{
        (a) Performance degradation occurs when jointly training I2I and T2I ReID tasks in a single model, compared to training each task independently.
        (b) The underlying cause lies in the semantic conflict: T2I emphasizes instance-specific attributes (e.g., ``holding a table") highlighted in text but ignored by I2I, despite shared identity-level cues such as clothing and gender.
    }
    \label{lblFigIntro}
\end{figure}

To achieve unified modeling for both I2I and T2I ReID, we attempt to train the two tasks within a single model jointly.
However, as shown in Figure~\ref{lblFigIntro}(a), this joint training strategy leads to a noticeable performance drop for both tasks compared to training them separately.
We attribute this degradation to the semantic conflict between the two tasks.
As illustrated in Figure~\ref{lblFigIntro}(b), while I2I and T2I share identity-level semantics (e.g., clothing, gender), T2I additionally relies on instance-specific attributes (e.g., actions or objects) described in natural language.
These attributes, such as the presence of a ``table", are often ignored by I2I alignment but emphasized in T2I supervision.
This inconsistency introduces conflicting optimization signals that negatively affect both tasks, ultimately impairing the effectiveness of joint training.

Prompt learning has recently achieved remarkable success in image-to-image (I2I) person re-identification (ReID) by injecting explicit identity-level semantics into the representation learning process \cite{I2I-CLIP1}. 
While identity-level prompts are also beneficial for text-to-image (T2I) ReID, directly extending prompt learning to this task remains non-trivial. Unlike I2I, which primarily relies on stable identity cues, T2I additionally requires fine-grained, instance-specific semantics—such as actions, gestures, or contextual objects—that are described in natural language but often vary across samples and views.
To bridge this gap, we draw inspiration from the prompt inversion mechanism proposed in GET~\cite{wang2025get}, which translates visual features into pseudo-textual prompts to facilitate cross-modal alignment. Building on this idea, we hypothesize that instance-level semantics can be directly inferred from either visual or textual inputs via modality-specific inversion networks.
Based on this insight, we formulate a unified hierarchical prompt structure: \textit{“A photo of [id-tokens] and [inst-tokens] person”}, where \textit{[id-tokens]} represent learnable identity-level semantics, and \textit{[inst-tokens]} encode dynamically generated instance-specific attributes. This formulation enables the model to jointly capture shared identity cues and task-specific semantic variations, thus accommodating the heterogeneous supervision signals of I2I and T2I ReID within a unified framework.

Motivated by this, we propose a unified ReID framework centered on Hierarchical Prompt Learning (HPL), which serves as the core mechanism to integrate identity-level stability with instance-level adaptability. HPL constructs structured prompt representations by combining the shared learnable identity prompt with dynamically generated instance prompt tokens, the latter derived from either visual or textual features via modality-specific inversion networks. These hierarchical prompts are injected into the CLIP backbone to provide task-aware semantic guidance for both I2I and T2I tasks.
To support effective prompt-driven learning, we introduce two complementary modules. First, the Task-Routed Transformer (TRT) extends the CLIP visual encoder with dual classification tokens, enabling the model to process both tasks within a shared backbone while preserving task-specific objectives. Second, the Cross-Modal Prompt Regularization (CMPR) module aligns instance-level prompt tokens generated from different modalities in the textual prompt space, enhancing semantic consistency and improving cross-modal generalization.
Together, these components form a cohesive architecture that enables unified modeling across modalities and supervision types, effectively bridging the gap between I2I and T2I ReID within a single framework.

The main contributions of this paper are summarized as follows:

\begin{itemize}
\item We propose a unified person ReID framework that jointly handles image-to-image (I2I) and text-to-image (T2I) retrieval tasks within a single architecture.

\item We design a task-aware prompting mechanism that integrates a Task-Routed Transformer (TRT) with a Hierarchical Prompt Learning (HPL) scheme. This allows the model to dynamically combine identity-level and instance-level semantics for both visual and textual queries.

\item We introduce a Cross-Modal Prompt Regularization (CMPR) strategy to align instance-level prompt tokens across modalities, enhancing semantic consistency. 

\item Extensive experiments on standard ReID benchmarks demonstrate the effectiveness of our framework, achieving state-of-the-art performance on both I2I and T2I tasks.
\end{itemize}

\section{Related Works}

\subsection{Image-to-Image Person Re-identification}

Image-to-Image Person Re-identification (I2I ReID) retrieves pedestrian images across camera views using a query image. Early methods employed CNNs for feature encoding and similarity matching \cite{I2I-Conv1,I2I-Conv2}. Recently, Transformer-based models \cite{ViT,I2I-ViT1,I2I-ViT2,I2I-ViT3,I2I-ViT4} have gained popularity due to their superior ability to capture global context.

More recently, the introduction of CLIP\cite{CLIP} has led to a surge in CLIP-based I2I ReID methods \cite{I2I-CLIP1, I2I-CLIP2, CLIP3DReID, MGMMPL, PIVL}. While CLIP is inherently designed for cross-modal alignment, standard I2I ReID datasets lack textual annotations. To utilize CLIP’s full potential, these works construct pseudo-text descriptions for images to assist training. For example, \cite{I2I-CLIP1} proposes a two-stage approach: prompt constructing and image-prompt alignment. Prompt-based methods have also been explored in text-\cite{Propot, PD} and clothes-changing-Re-ID\cite{MIPL}.
Building on this, \cite{I2I-CLIP2} introduces self-supervised strategies, such as masked text modeling and random image occlusion, to improve prompt quality.
And \cite{CLIP3DReID} further extracts body-shape-related phrases using CLIP.



\subsection{Text-to-Image Person Re-identification}

Text-to-Image Person Re-identification (T2I ReID) retrieves pedestrian images based on textual descriptions \cite{T2I1}. Early methods \cite{T2I-Early1} decompose both image and text into attribute components and conduct a fine-grained matching. Recent approaches propose end-to-end frameworks, but still follow the earlier idea of fine-grained alignment. Alignment strategies are typically explicit or implicit. Explicit methods, such as \cite{T2I-FineGrain4}, perform alignment on global and local features separately.
\cite{T2I-FineGrain5} addresses the asymmetry in cross-granularity alignment by introducing a non-symmetric alignment mechanism.
Implicit methods, such as \cite{RaSa,T2I-FineGrain3}, guide the model to learn fine-grained alignments through masked language modeling and similar techniques.

Recently, CLIP-based methods have gained prominence in the T2I ReID domain \cite{T2I-CLIP1,T2I-CLIP2,T2I-CLIP3,Harnessing}.
For example, \cite{T2I-CLIP1} introduces a multi-completeness constraint and dynamic attribute prompts to preserve CLIP’s generalization.
\cite{T2I-CLIP3} incorporates a feature filtering mechanism to mitigate the effects of occlusion.
Additionally, \cite{Harnessing} pretrains a vision-language model for T2I ReID using large-scale unlabelled pedestrian images paired with synthetic textual descriptions generated by multi-modal language models. More recent studies have also explored leveraging large vision-language models for person re-identification~\cite{ChatReID}.

However, most of these methods focus primarily on cross-modal alignment, while paying limited attention to intra-modal consistency. In contrast, our goal is to integrate I2I ReID with T2I ReID, thereby enabling an efficient multi-modal person retrieval system.


\begin{figure*}[h!]
\centering
\includegraphics[width=\linewidth]{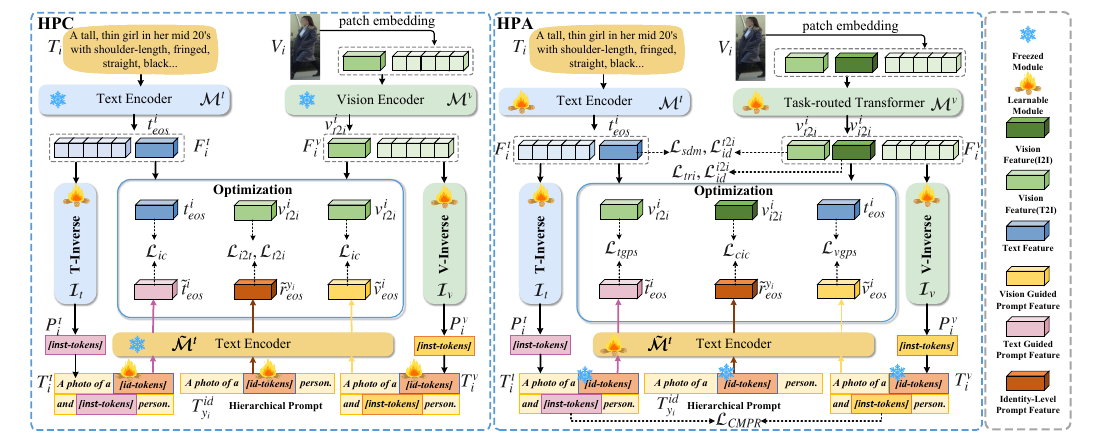}
\caption{
Overview of our framework. The framework comprises three core modules: 
(1) Task-Routed Transformer (TRT), which introduces dual classification tokens into the shared visual encoder to enable task-specific feature learning for I2I and T2I tasks; 
(2) Hierarchical Prompt Learning (HPL), which constructs and aligns identity-level and instance-level prompts via modality-specific inversion networks; 
(3) Cross-Modal Prompt Regularization (CMPR), which enforces semantic consistency between visual- and text-derived prompts at the instance level. 
Arrows indicate the directional flow of information and interactions among modules.
\label{lblFigMethod}
}
\end{figure*}

\section{Our Method}
We propose a unified framework named Hierarchical Prompt Learning, as shown in Figure~\ref{lblFigMethod}. It comprises three modules: a Task-Routed Transformer (TRT) with dual classification tokens for task-specific encoding, a Hierarchical Prompt Learning (HPL) module for hierarchical prompt generation and alignment, and a Cross-Modal Prompt Regularization (CMPR) module to enforce modality consistency via instance-level prompt supervision.

\subsection{Task-routed Transformer}

Previous methods typically use separate ViT backbones for T2I and I2I person ReID, despite the shared identity-relevant semantics (e.g., clothing, body structure) between tasks. The class token in ViTs naturally aggregates contextual information via self-attention. This mechanism implies that semantics aggregated into the class token are guided by the task-dependent objectives.
Motivated by this, we introduce a lightweight yet effective task-routed design by simply appending an additional class token into the CLIP visual encoder. Each class token is dedicated to one specific task: the original token $\feature{v}{i}{t2i}$ is optimized for T2I alignment, while the newly introduced token $\feature{v}{i}{i2i}$ focuses on identity discrimination in the I2I setting.

Specifically, given a visual input $V_{i} \in \mathbb{R}^{H \times W \times C}$ from either the T2I dataset $\mathcal{D}_{t2i}$ or the I2I dataset $\mathcal{D}_{i2i}$, we feed it into the CLIP visual encoder $\Encoder^v$, which is modified to include two class tokens. The resulting visual features are formulated as:
\begin{equation}
\Feature{v}{i} = [\feature{v}{i}{t2i}, \feature{v}{i}{i2i}, \feature{v}{i}{1}, \ldots, \feature{v}{i}{N}] = \Encoder^v(V_i),
\end{equation}
where $\feature{v}{i}{t2i}$ and $\feature{v}{i}{i2i}$ denote the classification tokens corresponding to the T2I and I2I tasks, respectively. The patch tokens $\feature{v}{i}{1}, \ldots, \feature{v}{i}{N}$ represent the spatial embeddings extracted from the image, where $N = H \times W / P^2$, and $P$ denotes the patch size.

To enable the two classification tokens to specialize in their respective tasks, we introduce a multi-objective supervision strategy that guides each token through task-relevant losses:
\begin{equation}
\Loss_{base} = \Loss_{sdm} + \Loss_{id}^{t2i} + \Loss_{tri} + \Loss_{id}^{i2i}
\end{equation}
Here, the T2I classification token $\feature{v}{i}{t2i}$ is optimized via the cross-modal similarity distribution matching loss $\Loss_{sdm}$ and the cross-modal identity classification loss $\Loss_{id}^{t2i}$\cite{T2I-CLIP2}, both of which utilize the paired text features for cross-modal alignment. In parallel, the I2I classification token $\feature{v}{i}{i2i}$ is supervised by an identity classification loss $\Loss_{id}^{i2i}$ and a triplet ranking loss $\Loss_{tri}$\cite{luobag}, promoting discriminative feature learning within the visual modality.

\subsection{Hierarchical Prompt Learning}

While TRT enables shared visual backbones, it lacks fine-grained semantic modeling. I2I benefits from identity-level prompts, as shown in CLIP-ReID~\cite{I2I-CLIP1}, but T2I further relies on instance-specific details (e.g., clothing color, accessories). To address this, we propose Hierarchical Prompt Learning (HPL), which combines identity- and instance-level prompts to guide both tasks. HPL includes two modules: \textit{ Hierarchical Prompt Construction} and \textit{Hierarchical Prompt Alignment}.

\textbf{Hierarchical Prompt Construction.}
To provide unified textual supervision for both I2I and T2I person ReID, we design a hierarchical language template $T_i$:
\textit{“A photo of [id-tokens] and [inst-tokens] person”},
which differs from conventional prompt designs (e.g., CLIP-ReID). Here, the identity-level component \textit{[id-tokens]} consists of a fixed number of learnable tokens, while the instance-level component \textit{[inst-tokens]} is instantiated by pseudo-text tokens converted from either image or text features. This design injects both identity-level and instance-specific semantics into the prompt.

To enrich the prompt with fine-grained instance cues, we introduce two modality-specific inverse networks $\mathcal{I}_v$ and $\mathcal{I}_t$ to generate pseudo-text tokens from vision and text features, respectively. Given an image $V_i$ and its associated prompt $T_i$, we compute the visual and textual features:
\begin{equation}
\Feature{v}{i} = \Encoder^v(V_i), \quad \Feature{t}{i} = \Encoder^t(T_i),
\end{equation}
which are then fed into the corresponding inverse networks:
\begin{equation}
\Prm{t}{i} = \mathcal{I}_t(\Feature{t}{i}), \quad \Prm{v}{i} = \mathcal{I}_v(\Feature{v}{i}),
\end{equation}
where $\mathcal{I}_v$ and $\mathcal{I}_t$ are composed of $N_{\mathcal{I}}$ Transformer layers following the ViT block architecture.
These pseudo-text tokens are inserted into the hierarchical template to form: $T^v_i$: “A photo of [id-tokens] and $\Prm{v}{i}$ person.”;
$T^t_i$: “A photo of [id-tokens] and $\Prm{t}{i}$ person.” The resulting hierarchical prompts are then encoded via another CLIP text encoder $\tilde{\Encoder}^t$:
\begin{equation}
\featureT{v}{i}{eos} = \tilde{\Encoder}^t(T^v_i), \quad \featureT{t}{i}{eos} = \tilde{\Encoder}^t(T^t_i).
\end{equation}
To ensure that these pseudo prompts faithfully retain modality-specific semantics, we apply an inversion consistency loss:
\begin{equation}
\Loss_{ic} = \frac{1}{|B|} \sum_{i \in B} ||\featureT{v}{i}{eos} - \feature{v}{i}{t2i}||_2^2 + \frac{1}{|B_{t2i}|} \sum_{i \in B_{t2i}} ||\featureT{t}{i}{eos} - \feature{t}{i}{eos}||_2^2,
\end{equation}
where $B_{t2i}$ and $B$ indicate the batch for T2I and the entire batch, respectively, and $||\cdot||_2^2$ denotes the squared L2 norm.
This supervision encourages the reconstructed prompts to reflect their input semantics while enhancing cross-modal transferability. During this stage, the encoders $\Encoder^v$, $\Encoder^t$ and $\tilde{\Encoder}^t$ are frozen, and only the inverse networks and prompts are updated.

To encode identity-level semantics, we instantiate a simplified version of the hierarchical prompt:
$T^{id}_{y_i}$: \textit{“A photo of [id-tokens] person”},
where \textit{[id-tokens]} corresponds to the shared learnable identity-level tokens. This template is also passed the text encoder $\tilde{\Encoder}^t$ to obtain the reference text representation:
\begin{equation}
\featureT{r}{y_i}{eos} = \tilde{\Encoder}^t(T^{id}_{y_i}).
\end{equation}
To guide identity-level alignment, we adopt two contrastive losses:
\begin{equation}
    \Loss_{t2i} = -\frac{1}{|B|}\sum\limits_{i\in B}\log\frac{\exp[sim(\feature{v}{i}{t2i}, \featureT{r}{y_i}{eos})]}{\sum\limits_{j\in B} \exp[sim(\feature{v}{j}{t2i}, \featureT{r}{y_i}{eos})]},
\end{equation}
\begin{equation}
    \Loss_{i2t} = -\frac{1}{|B|}\sum\limits_{i\in B}\log\frac{\exp[sim(\feature{v}{i}{t2i}, \featureT{r}{y_i}{eos})]}{\sum\limits_{j\in B} \exp[sim(\feature{v}{i}{t2i}, \featureT{r}{y_j}{eos})]},
\end{equation}
where $sim(\cdot, \cdot)$ denotes cosine similarity. These losses encourage the identity-level prompt to serve as a reliable semantic anchor during downstream training.

\textbf{Hierarchical Prompt Alignment.} 
In Hierarchical Prompt Alignment, we aim to utilize hierarchical prompts from Hierarchical Prompt Construction to help the model learn the necessary semantic features for T2I and I2I tasks.
For the T2I task, we utilize the full hierarchical prompt—comprising both identity-level and instance-level components—to guide the visual and textual representations, respectively. Specifically, the visual classification token $\feature{v}{i}{t2i}$ is encouraged to align with the textual embedding $\featureT{t}{i}{eos}$ derived from the pseudo-prompt $T^{v}_{i}$, while the textual feature $\feature{t}{i}{eos}$ is aligned with $\featureT{v}{i}{eos}$ generated from the pseudo-prompt $T^{t}_{i}$. This is enforced through the instance-level prompt alignment loss $\Loss_{ILPA}$, which consists of text- and vision-guided prompts supervised loss $\Loss_{tgps}$ and $\Loss_{vgps}$:

\begin{equation}
    \Loss_{tgps}=\frac{1}{|B_{t2i}|}\sum\limits_{i\in B_{t2i}}
    ||\featureT{t}{i}{eos}-\feature{v}{i}{t2i}||_2^2 \nonumber
\end{equation}
\begin{equation}
    \Loss_{vgps}=
    \frac{1}{|B_{t2i}|}\sum\limits_{i\in B_{t2i}}
    ||\featureT{v}{i}{eos}-\feature{t}{i}{eos}||_2^2 \nonumber
\end{equation}
\begin{equation}
\Loss_{ILPA}=\Loss_{tgps}+\Loss_{vgps}
\end{equation}
For the I2I task, given a pedestrian image, we use the identity-level prompt $T^{id}_{y_i}$, which is encoded by the text encoder $\tilde{\Encoder}^{t}$ to produce the textual representation $\featureT{r}{y_i}{eos}$. We then apply a cross-modal identity classification loss $\Loss_{cic}$ to supervise the alignment between the image feature $\feature{v}{i}{i2i}$ and the identity prompt:
\begin{equation}
\Loss_{cic} = -\sum\limits_{i\in B} \log\frac{\exp[sim(\feature{v}{i}{i2i}, \featureT{r}{y_i}{eos})]}{\sum\limits_{j=1}^{N_{id}}\exp[sim(\feature{v}{i}{i2i}, \featureT{r}{j}{eos})]}
\end{equation}
where $N_{id}$ denotes the total number of identities in the training set.

\textbf{Advantages of HPL.}
Compared to previous prompt learning strategies that typically focus only on identity-level templates or rely on fixed textual embeddings, our hierarchical prompt scheme brings three key advantages:
(1) \textit{Fine-grained Adaptability:} The instance-level pseudo tokens capture unique instance-specific attributes (e.g., outfit, accessories), enabling the model to attend to nuanced visual cues beyond class identity.
(2) \textit{Bidirectional Cross-Modal Alignment:} By decoding visual/textual features into pseudo-text and enforcing both text-to-image and image-to-text alignment, our approach bridges modality discrepancies and improves semantic coherence across modalities.
(3) \textit{Prompt-based Harmonization:} Since the instance-level and identity-level prompts are concatenated together, the ILPA loss not only enhances the alignment of instance-specific semantics across modalities but also preserves identity consistency.
These strengths allow the proposed HPL framework to generalize better across diverse ReID tasks, particularly under the challenging T2I scenario where textual guidance plays a crucial role.

\subsection{Cross-Modal Prompt Regularization}

While HPL captures both identity- and instance-level semantics, the alignment between visual and textual prompts remains implicit. The instance-level prompts $\Prm{v}{i}$ and $\Prm{t}{i}$ may encode modality-specific biases, causing semantic inconsistency—especially detrimental for T2I grounding. To mitigate this, we introduce cross-modal prompt regularization, which aligns pseudo-text tokens in the textual token space via inverse networks $\mathcal{I}_v$ and $\mathcal{I}_t$ directly.
Formally, given an image-text pair $(V_i, T_i)$, we extract the corresponding instance-level prompts as $\Prm{v}{i} = \mathcal{I}_v(\Feature{v}{i})$ and $\Prm{t}{i} = \mathcal{I}_t(\Feature{t}{i})$, where $\Feature{v}{i} = \Encoder^v(V_i)$ and $\Feature{t}{i} = \Encoder^t(T_i)$ denote the visual and textual features, respectively. Then, the prompt-level alignment loss is defined as:
\begin{equation}
\Loss_{CMPR} = \frac{1}{|B_{t2i}|} \sum\limits_{i \in B_{t2i}} \|\Prm{t}{i} - \Prm{v}{i}\|_F^2,
\end{equation}
where $||\cdot||_F^2$ denotes the Frobenius norm, and $B_{t2i}$ represents the batch of T2I training samples.


This regularization promotes a unified prompt representation for both modalities, ensuring image-guided prompts convey meaningful semantics that align with text-guided prompts, reducing cross-modal semantic drift and improving text-guided visual retrieval.

\subsection{Optimization}

The training process consists of two stages: constructing hierarchical prompts and leveraging them for representation learning.

\textbf{Stage I: Prompt Construction.}
In this stage, we focus on learning discriminative and semantically consistent prompts. The contrastive losses $\Loss_{t2i}$ and $\Loss_{i2t}$ are applied to guide identity-level prompt learning, while the inversion consistency loss $\Loss_{ic}$ ensures that the instance-level pseudo-prompts remain aligned with the original visual and textual features. The overall objective in this stage is:
\begin{equation}
    \Loss_{construct} = \Loss_{t2i} + \Loss_{i2t} + \Loss_{ic}.
\end{equation}

\textbf{Stage II: Representation Learning.}
With the constructed prompts, we proceed to optimize cross-modal feature learning. The base loss $\Loss_{\text{base}}$ supervises the I2I and T2I tasks. To enhance semantic alignment, we introduce identity-level and instance-level prompt alignment losses, $\Loss_{cic}$ and $\Loss_{ILPA}$. Additionally, we incorporate the cross-modal prompt regularization loss $\Loss_{CMPR}$ to further encourage consistency between visual- and text-derived prompts. The total loss for this stage is:
\begin{equation}
    \Loss_{total} = \Loss_{base} + \Loss_{cic} + \lambda_1 \Loss_{ILPA} + \lambda_2 \Loss_{CMPR}.
\end{equation}

This two-stage optimization facilitates joint learning of multi-level prompt representations and cross-modal features, effectively bridging the modality gap while preserving fine-grained identity semantics.

\begin{table*}[t]\small
    \centering
        \caption{
    	Comparison with T2I ReID methods on CUHK-PEDES, ICFG-PEDES and RSTPReID datasets. Experiments are conducted on combinations of T2I and I2I datasets. Specifically, the three T2I datasets in this table are respectively paired with the three I2I datasets used in Table~\ref{lblTblCompI2I}.} \vspace{-2mm}
    \begin{tabular}{c|c|ccccccccccc}
        \toprule  
          \multirow{2}{*}{\textbf{Method}}       & \multirow{2}{*}{\textbf{Reference}} &  \multicolumn{2}{c}{\textbf{CUHK-PEDES}} & \multicolumn{2}{c}{\textbf{ICFG-PEDES}} & \multicolumn{2}{c}{\textbf{RSTPReID}} \\ 
              &    & \textbf{Rank-1} & \textbf{mAP} & \textbf{Rank-1} & \textbf{mAP} & \textbf{Rank-1} & \textbf{mAP} \\
        \midrule 
        CFine     \cite{CFine}     & TIP'23  & 69.57 & -     & 60.83 & -     & 50.55 & -     \\
        IRRA      \cite{T2I-CLIP2} & CVPR'23 & 73.38 & 66.13 & 63.46 & 38.06 & 60.20 & 47.17 \\
        MDRL      \cite{MDRL}      & AAAI'24 & 74.56 & -     & 65.88 & -     & -     & -     \\
        TBPS-CLIP \cite{TBPS-CLIP} & AAAI'24 & 73.54 & 65.38 & 65.05 & 39.83 & 61.95 & 48.26 \\
        IRLT      \cite{IRLT}      & AAAI'24 & 74.46 & -     & 64.72 & -     & 61.49 & -     \\
        UMSA      \cite{UMSA}      & AAAI'24 & 74.25 & 66.15 & 65.62 & 38.78 & 63.40 & 49.28 \\
        LSPM      \cite{LSPM}      & TMM'24  & 74.38 & 67.74 & 64.40 & 42.60 & -     & -     \\
        FSRL      \cite{FSRL}      & ICMR'24 & 74.86 & 67.57 & 64.93 & 40.67 & 60.65 & 48.18 \\
        Propot    \cite{Propot}    & MM'24   & 74.89 & 67.12 & 65.12 & 42.93 & 61.87 & 47.82 \\
        MMRef     \cite{MMRef}     & TMM'25  & 72.25 & -     & 63.50 & -     & 56.20 & -     \\
        \midrule 
        HPL(ours)        & This Paper & \textbf{76.28} & \textbf{70.90} & \textbf{66.61} & \textbf{44.14} & \textbf{64.00} & \textbf{53.13} \\
        \bottomrule
    \end{tabular}
    \label{lblTblCompT2I}
\end{table*}

\begin{table*}\small
    \centering
        \caption{
    	Comparison with I2I ReID methods on Market1501, MSMT17 and DukeMTMC datasets. The three I2I datasets in this table are respectively paired with the three T2I datasets used in Table~\ref{lblTblCompT2I}.} \vspace{-2mm}
    \begin{tabular}{c|c|ccccccccccc}
        \toprule  
        \multirow{2}{*}{\textbf{Method}} & \multirow{2}{*}{\textbf{Reference}} &  \multicolumn{2}{c}{\textbf{Market1501}} & \multicolumn{2}{c}{\textbf{MSMT17}} & \multicolumn{2}{c}{\textbf{DukeMTMC}} \\ 
                 &  & \textbf{Rank-1} & \textbf{mAP} & \textbf{Rank-1} & \textbf{mAP} & \textbf{Rank-1} & \textbf{mAP} \\
        \midrule 
        CDNet      \cite{CDNet}      & CVPR'21 & 95.10  & 86.00  & 78.90 & 54.70 & 88.60           & 76.80 \\
        TransReID  \cite{TransReID}  & ICCV'21 & 95.20  & 88.90  & 85.30 & 67.40 & \textbf{90.70} & 82.00 \\
        DRL-Net    \cite{DRL-Net}    & TMM'22  & 94.70  & 86.90  & 78.40 & 55.30 & 88.10          & 76.60 \\
        DCAL       \cite{DCAL}       & CVPR'22 & 94.70  & 87.50  & 83.10 & 64.00 & 89.00          & 80.10 \\
        CLIP-ReID  \cite{I2I-CLIP1}  & AAAI'23 & 95.50  & 89.60  & 88.70 & 73.40 & 90.00          & 82.50 \\
        CLIP3DReID \cite{CLIP3DReID} & CVPR'24 & 95.60  & 88.40  & 81.50 & 61.20 & -              & -     \\
        \midrule 
        HPL(ours)         & This Paper & \textbf{95.99}  & \textbf{89.82}  & \textbf{91.04} & \textbf{79.01} & 90.35 & \textbf{82.93} \\
        \bottomrule
    \end{tabular}
    \label{lblTblCompI2I}
\end{table*}
\section{Experiments}

\subsection{Datasets and Evaluation Protocols}
We evaluate our method on both image-to-image (I2I) and text-to-image (T2I) person re-identification tasks using three dataset combinations: \textit{CUHK-PEDES + Market1501}, \textit{ICFG-PEDES + MSMT17}, \textit{RSTPReid + DukeMTMC-ReID}.
CUHK-PEDES~\cite{T2I1} is a large-scale text-to-image benchmark, containing 40,208 images of 13,003 identities, each image annotated with two textual descriptions.
ICFG-PEDES~\cite{ICFG-PEDES} contains 54,522 images from 4,102 identities, with each image paired with one natural language sentence.
RSTPReid~\cite{RSTP-ReID} consists of 20,505 images from 4,101 identities captured by 15 surveillance cameras.
Market1501~\cite{Market1501} is a widely used ReID benchmark, providing 32,668 images of 1,501 identities collected from 6 cameras.
MSMT17~\cite{MSMT} contains 126,441 images of 4,101 identities under various lighting conditions and viewpoints from 15 cameras.
DukeMTMC-ReID~\cite{DukeMTMC-ReID} consists of 36,411 images with 1,404 identities captured across 8 cameras.
Following common protocols, we report Rank-1 accuracy and mean Average Precision (mAP) for all tasks.

\begin{table}
    \centering
        \caption{Impact of TRT, HPL, and CMPR modules on Rank-1 and mAP on the CUHK-PEDES + Market1501 dataset. }\vspace{-2mm}
    \resizebox{0.45 \textwidth}{!}{
        \begin{tabular}{ccc|cc|cc}
            \toprule 
               \multicolumn{3}{c}{\textbf{Components}} & \multicolumn{2}{c}{\textbf{T2I}} & \multicolumn{2}{c}{\textbf{I2I}} \\
            \textbf{TRT} & \textbf{HPL} & \textbf{CMPR}   & \textbf{Rank-1} & \textbf{mAP} & \textbf{Rank-1} & \textbf{mAP} \\
            \midrule 
              \ding{55}          & \ding{55}           & \ding{55}           & 74.22 & 70.45 & 94.50 & 86.91  \\
            \checkmark  & \ding{55}          &\ding{55}            & 75.27 & 70.80 & 95.36 & 88.98  \\
            \checkmark  & \checkmark &  \ding{55}          & 75.60 & 70.88 & 95.57 & 89.72  \\
            \checkmark  & \checkmark & \checkmark & \textbf{76.28} & \textbf{70.89} & \textbf{95.99} & \textbf{89.82}  \\
            \bottomrule
        \end{tabular} 
    }

    \label{lblTabAblation1}
\end{table}

\subsection{Implementations}
We adopt CLIP pretrained on LUPerson and large-scale synthetic image-text pairs as the backbone. Input images are resized to $384 \times 128$, and textual inputs are truncated to 77 tokens.
During the HPC stage, we train the model for 10 epochs using the Adam optimizer. The learning rates for the decoder and identity-level prompts are set to $5 \times 10^{-5}$ and $0.02$, respectively, with an exponential decay factor of 0.8 per epoch. Both the visual and textual inverse networks are configured with $N_{\mathcal{I}} = 4$ transformer layers.
In the HPA stage, training proceeds for 60 epochs. A linear warm-up is applied in the first 5 epochs (from $10^{-6}$ to $10^{-5}$), followed by cosine annealing. Each batch contains 64 image-text pairs from the T2I dataset and 64 images from the I2I dataset, with 4 instances per identity.
To avoid data leakage and identity confusion, we remove test samples from training splits and unify identity labels across datasets. This ensures that samples belonging to the same person in different datasets are treated consistently during training.
The loss weights $\lambda_1$ and $\lambda_2$ are set to 0.4 and 0.06, respectively. All experiments are conducted on a single NVIDIA RTX 4090 GPU.

\subsection{Comparison with State-of-the-Art Methods}
We report the performance of our unified framework on both T2I and I2I person Re-ID tasks across six benchmark datasets.

\textbf{T2I ReID.}
We evaluated our method on several T2I ReID datasets. As shown in Table~\ref{lblTblCompT2I}, our method achieves competitive performance across all popular datasets. Specifically, our method obtains Rank-1 accuracy of 76.28\%, 66.61\%, and 64.00\% on the CUHK-PEDES, ICFG-PEDES, and RSTPReID datasets, respectively.
Moreover, unlike most existing methods, our approach simultaneously achieves strong performance on the I2I ReID task, highlighting its superiority in multi-task person retrieval.

\begin{table}
    \centering
        \caption{Ablation of instance-level alignment using vision-guided, text-guided, and dual-modality prompt supervision on CUHK-PEDES + Market1501.} \vspace{-2mm}
    \resizebox{0.45 \textwidth}{!}{
        \begin{tabular}{cc|cc|cc}
            \toprule 
               \multicolumn{2}{c}{\textbf{ILPA}} & \multicolumn{2}{c}{\textbf{T2I}} & \multicolumn{2}{c}{\textbf{I2I}} \\
            $\Loss_{tgps}$ & $\Loss_{vgps}$ & \textbf{Rank-1} & \textbf{mAP} & \textbf{Rank-1} & \textbf{mAP} \\
            \midrule 
                \ding{55}        &\ding{55}             & 75.27 & 70.80 & 95.36 & 88.98  \\
            \checkmark  & \ding{55}            & 75.58 & 70.84 & 95.42 & 89.35  \\
               \ding{55}         & \checkmark  & 75.60 & 70.82 & \textbf{95.69} & 89.59  \\
            \checkmark  & \checkmark  & \textbf{75.60} & \textbf{70.88} & 95.57 & \textbf{89.72}  \\
            \bottomrule
        \end{tabular}
    }

    \label{lblTabAblation2}
\end{table}
\textbf{I2I ReID.}
We also evaluated our method on several widely used I2I ReID benchmarks. As shown in Table~\ref{lblTblCompI2I}, our approach outperforms CLIPReID \cite{I2I-CLIP1}, a state-of-the-art method that also leverages learnable prompts for training.
Specifically, our method achieves Rank-1 accuracies of 95.99\%, 91.04\%, and 90.35\% on Market1501, MSMT17, and DukeMTMC-ReID datasets, respectively.
The corresponding mAP scores reach 89.82\%, 79.01\%, and 82.93\%.

\subsection{Ablation Studies and Analysis}

\begin{figure}[t]
    \centering
    \includegraphics[width=\linewidth]{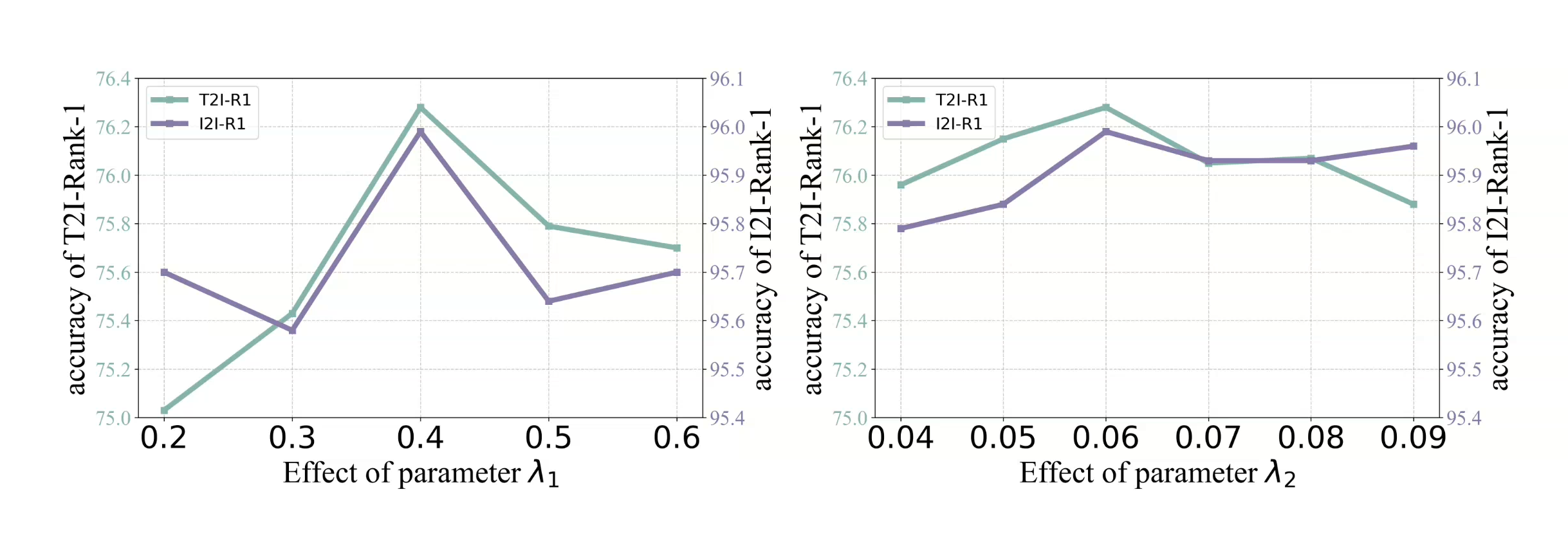}
    \label{fig:side:a}
    \caption{
Effect of $\lambda_1$ and $\lambda_2$ on Rank-1 accuracy on CUHK-PEDES + Market1501.
    }
    \label{lblFigHyperParam}
\end{figure}

To verify the effectiveness of each component of our method, we conducted ablation experiments on the CUHK-PEDES + Market1501 dataset combination. The results demonstrate that our method successfully mitigates the conflict between the two tasks, enabling better integration of I2I and T2I ReID.

\textbf{Effectiveness of TRT.}
To evaluate the effectiveness of the Task-routed Transformer, we compared it with a baseline that uses a single class token. As shown in Table~\ref{lblTabAblation1}, introducing a dual class token design leads to a 1.07\% improvement in T2I Rank-1 accuracy and a 2.07\% increase in I2I mAP.
These findings also suggest that T2I ReID and I2I ReID rely on different types of information. Using a shared feature representation for both tasks may result in mutual interference, whereas the dual-token structure helps decouple task-specific semantics.

\textbf{Effectiveness of HPL.}
To evaluate the effectiveness of our Hierarchical Prompt Learning (HPL) module, we conduct ablation studies based on TRT with and without HPL. As shown in Table~\ref{lblTabAblation1}, introducing HPL yields considerable performance gains for both I2I and T2I tasks. This indicates that identity-level and sample-level prompts help the model capture more refined and task-aware features. However, the improvement remains limited when HPL is used alone.

We further investigate using uni-modal sample-level prompts from either the visual or textual modality.
As shown in Table~\ref{lblTabAblation2}, both achieve gains over the baseline, confirming the benefit of sample-level guidance.
Compared to uni-modal prompts, the dual-modal setting brings a slight drop in I2I Rank-1 (-0.12\%) but improves mAP (+0.13\%), suggesting more comprehensive identity coverage.
The impact on T2I is marginal, likely due to sufficient grounding from the paired text.

\textbf{Effectiveness of CMPR.}
Based on TRT and HPL, we conduct ablation studies with and without Cross-Modal Prompt Regularization (CMPR).
As shown in Table~\ref{lblTabAblation1}, adding CMPR together with HPL leads to significant performance improvements across both I2I and T2I tasks(+1.01\% on T2I Rank1 and + 0.84\% on I2I mAP).
This demonstrates that CMPR serves as an effective regularizer that reinforces prompt-level consistency and promotes better semantic alignment between modalities.

\begin{figure}[t]
    \includegraphics[width=0.95\linewidth]{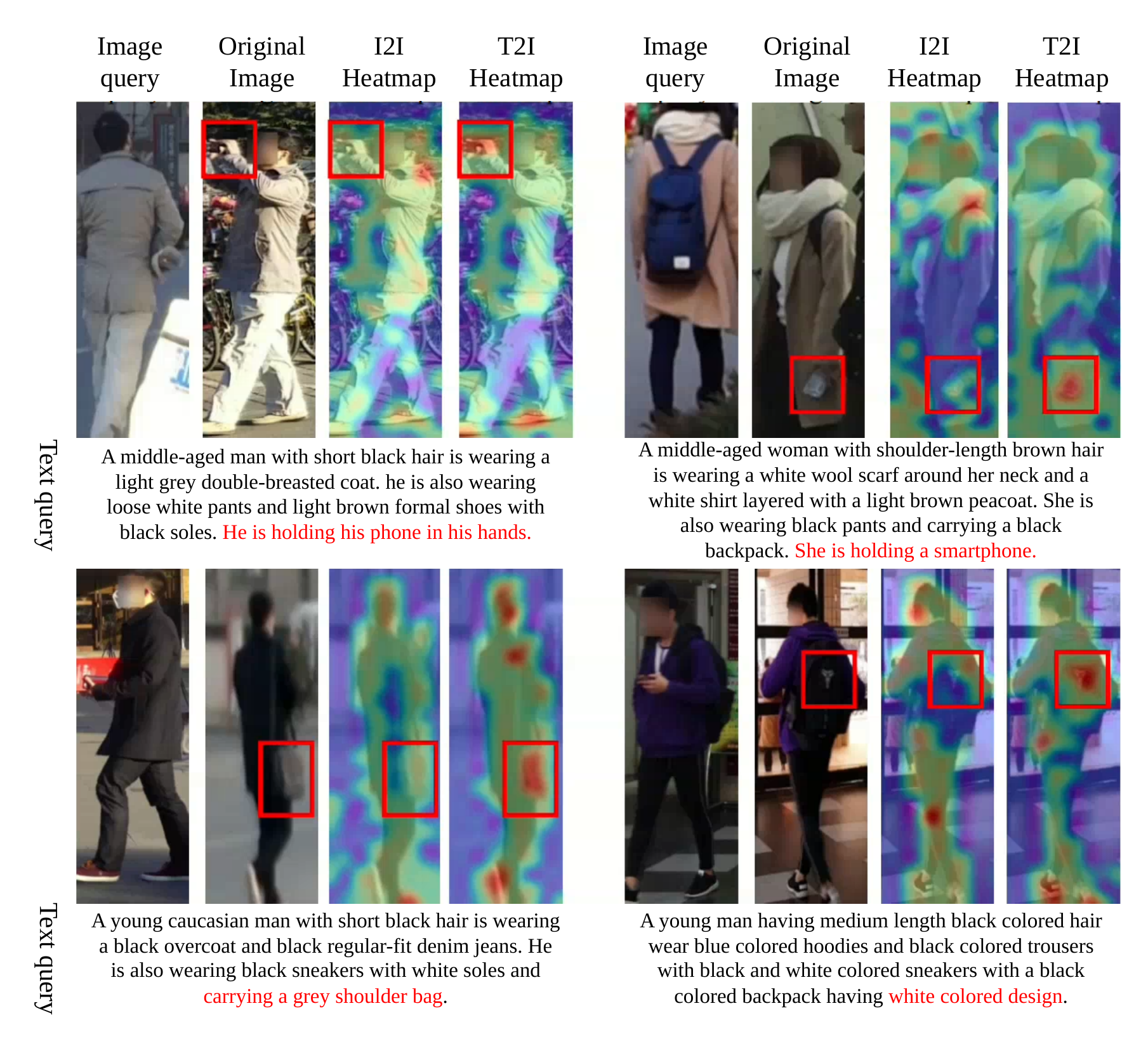}
    \caption{
Grad-CAM comparison of I2I and T2I tokens, showing stronger focus on view-specific cues in T2I aligned with text descriptions.
    }
    \label{lblFigHeatmap}
\end{figure}

\textbf{Hyperparameter Analysis.}
To investigate the effect of different values of $\lambda_1$ and $\lambda_2$ on model performance,
we conducted a sensitivity analysis on the CUHK-PEDES + Market1501 dataset combination.
As shown in Figure~\ref{lblFigHyperParam}, both excessively small and large values of $\lambda_1$ and $\lambda_2$ lead to performance degradation.
Based on experimental results, we set $\lambda_1=0.4$ and $\lambda_2=0.06$ to achieve optimal results.

\textbf{Visualization Results.}
We visualize attention maps using Grad-CAM~\cite{GradCam} to analyze the behavior of I2I and T2I classification tokens (Figure~\ref{lblFigHeatmap}). I2I tokens mainly focus on identity-related regions such as clothing and body shape, which remain consistent across views. In contrast, T2I tokens emphasize instance-specific details mentioned in text, e.g., phones or shoulder bags. For instance, in the second row, T2I attention highlights the grey shoulder bag while I2I ignores it. This modality-specific focus shows the model’s adaptive disentanglement of task-relevant semantics, alleviating supervision conflicts in joint training.

\section{Conclusion}

In this paper, we present a unified framework for image- and text-based person re-identification, jointly optimizing both tasks via task-aware prompt learning. A Task-Routed Transformer with dual classification tokens enables task-specific representation within a shared encoder. To capture multi-level semantics, a Hierarchical Prompt Learning module disentangles identity- and instance-level cues through modality-specific inversion. Furthermore, Cross-Modal Prompt Regularization aligns pseudo-prompts across modalities to reduce semantic inconsistency. Experiments on six benchmarks show that our method consistently surpasses state-of-the-art approaches on both I2I and T2I tasks.

\section{Acknowledgments}
This work was supported in part by the National Science Foundation of China under Grant 62276120 and Grant 61966021, the Yunnan Fundamental Research Projects under Grant 202301AV070004 and Grant 202401AS070106, the Major Science and Technology Special Projects of Yunnan Province under Grant 202502AD080006.

\bibliography{aaai2026}

\end{document}